\pdfoutput=1

\documentclass[11pt]{article}

\usepackage[final]{acl}

\usepackage{times}
\usepackage{latexsym}
\usepackage{amsmath}
\usepackage{amssymb}
\usepackage{multirow}
\usepackage{booktabs}
\usepackage{float}
\usepackage{afterpage}

\usepackage[T1]{fontenc}

\usepackage[utf8]{inputenc}

\usepackage{microtype}

\usepackage{inconsolata}

\usepackage{graphicx}

\title{Boosting LLM’s Molecular Structure Elucidation with\\ Knowledge Enhanced Tree Search Reasoning}

\author{
 \textbf{Xiang Zhuang\textsuperscript{1,2}},
 \textbf{Bin Wu\textsuperscript{3}},
 \textbf{Jiyu Cui\textsuperscript{1}},
 \textbf{Kehua Feng\textsuperscript{1,2}}, \\
 \textbf{Xiaotong Li\textsuperscript{1,2}}, 
 \textbf{Huabin Xing\textsuperscript{1,2}},
 \textbf{Keyan Ding\textsuperscript{2}},
 \textbf{Qiang Zhang\textsuperscript{1,2}\footnotemark[1]},
 \textbf{Huajun Chen\textsuperscript{1,2}\thanks{Corresponding author.}}
\\
\\
\textsuperscript{1} Zhejiang University \\
\textsuperscript{2} ZJU-Hangzhou Global Scientific and Technological Innovation Center\\
\textsuperscript{3} University College London
\\
\texttt{\{zhuangxiang, qiang.zhang.cs, huajunsir\}@zju.edu.cn}
}

\begin{document}
\maketitle
\begin{abstract}
Molecular structure elucidation involves deducing a molecule's structure from various types of spectral data, which is crucial in chemical experimental analysis. While large language models (LLMs) have shown remarkable proficiency in analyzing and reasoning through complex tasks, they still encounter substantial challenges in molecular structure elucidation.  We identify that these challenges largely stem from LLMs' limited grasp of specialized chemical knowledge. In this work, we introduce a \textbf{K}nowledge-enhanced reasoning framework for \textbf{M}olecular \textbf{S}tructure \textbf{E}lucidation (K-MSE), leveraging Monte Carlo Tree Search for test-time scaling as a plugin. Specifically, we construct an external molecular substructure knowledge base to extend the LLMs' coverage of the chemical structure space. Furthermore, we design a specialized molecule-spectrum scorer to act as a reward model for the reasoning process, addressing the issue of inaccurate solution evaluation in LLMs. Experimental results show that our approach significantly boosts performance, particularly gaining more than 20\% improvement on both GPT-4o-mini and GPT-4o\footnote{Code repository: \url{https://github.com/HICAI-ZJU/K-MSE}.}.
\end{abstract}

\section{Introduction}
Recent studies have demonstrated the broad application potential of large language models (LLMs)~\cite{DBLP:journals/corr/abs-2303-08774,dubey2024llama} in chemistry-related tasks, such as reaction prediction~\cite{m2024augmenting} and molecule generation~\cite{DBLP:conf/iclr/LiuWYW0GX24}. However, despite progress in these fields, fully leveraging the deep thinking and reasoning capabilities of LLMs to address complex problems in chemistry, such as molecular structure elucidation, remains a significant challenge. Molecular structure elucidation is a fundamental task in chemical experimental analysis, involving the deduction of molecular structures from various types of spectral data,  such as nuclear magnetic resonance (NMR), infrared (IR) spectroscopy, etc.~\cite{guo2024can}. Accurate elucidation of molecular structures is essential for chemical research, as it forms a critical step in interpreting experimental results~\cite{xue2023advances}. Nevertheless, this task is inherently complex, often requiring significant time and expertise. Even skilled professionals typically need 10 to 15 minutes to reason through the structure of a single molecule~\cite{field2020organic}.  Therefore, automating the interpretation of spectra and accurately deducing molecular structures using LLMs has the potential to greatly enhance experimental efficiency and drive the automation of chemical research~\cite{dai2024autonomous}.

Recent studies have demonstrated that increasing the number of reasoning steps, especially with tree search~\cite{DBLP:conf/emnlp/HaoGMHWWH23,DBLP:journals/corr/abs-2406-07394,DBLP:journals/corr/abs-2410-02884}, can greatly improve LLMs' ability to solve complex problems. This highlights the importance of test-time scaling laws~\cite{DBLP:journals/corr/abs-2408-03314,openaio1}.
However, existing methods are not easily applicable to molecular structure elucidation. This is largely due to LLMs' limited grasp of chemical molecular knowledge~\cite{LKM}, which can be attributed to two key factors.

\begin{figure}[t]
    \centering
    \includegraphics[width=\linewidth]{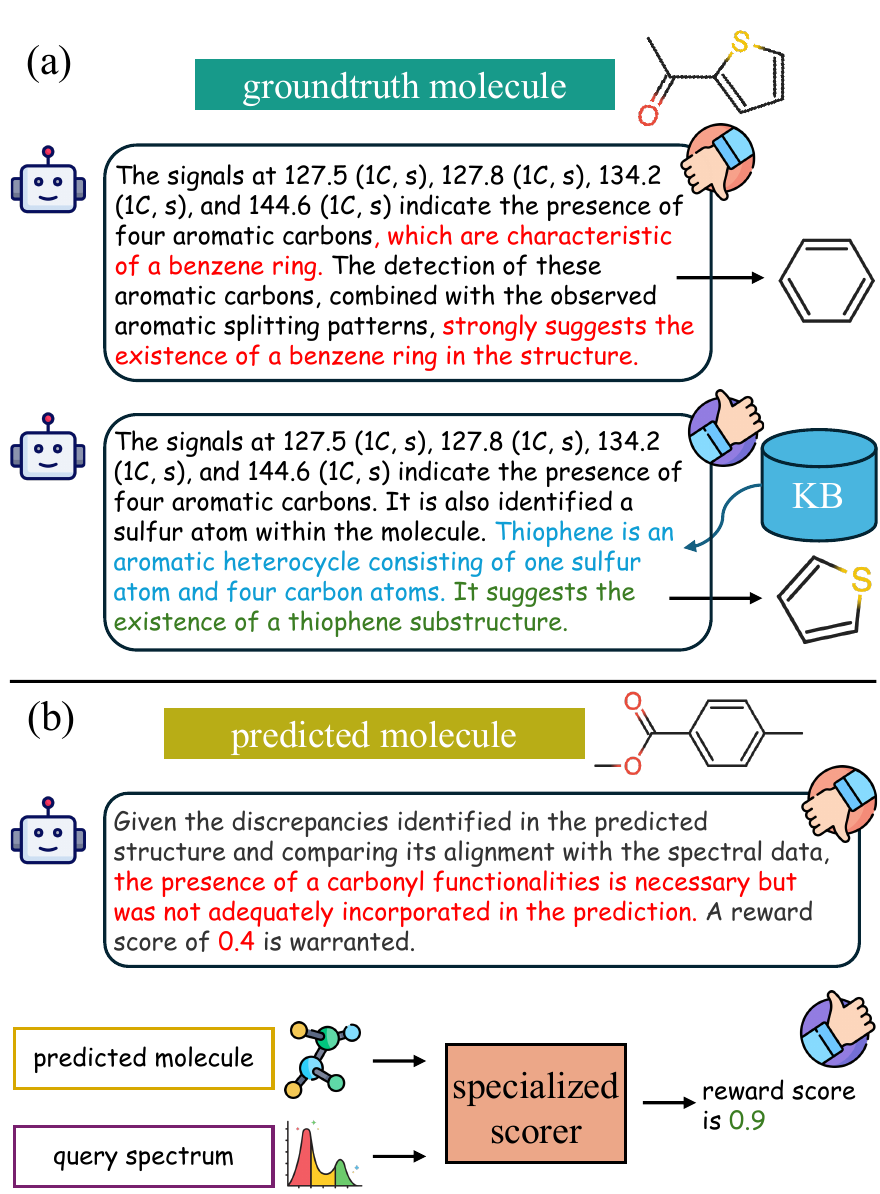} 
    \caption{(a) By incorporating a knowledge base, we enhance the ability of LLMs to handle a broader range of chemical molecular structures. (b) Through the development of a specialized scorer, we enable accurate evaluation of potential solutions.} 
    \label{fig:intro} 
\end{figure}

First, LLMs lack comprehensive coverage of chemical molecular structure space. As shown in Figure~\ref{fig:intro}(a), for instance, thiophene—an aromatic heterocycle composed of one sulfur atom and four carbon atoms—is a structure that LLMs often struggle to analyze accurately. Although the model may recognize that the structure contains at most four aromatic carbon atoms and one sulfur atom, it frequently misidentifies it as the benzene ring (the most common aromatic substructure with six carbon atoms), overlooking the limitation on the number of aromatic carbon atoms.
This suggests that LLMs lack critical substructure knowledge, hindering their ability to fully comprehend and accurately infer complex and unusual chemical structures.

{Second, LLMs cannot accurately evaluate and correct their reasoning process.
This capability is crucial for tree search-based reasoning, as it guides reasoning to be more effective and efficient~\cite{DBLP:conf/icml/WanFWM00024}. 
Timely evaluation and correction provide the necessary feedback, helping LLMs identify potential issues and optimize accordingly. However, we have observed that in molecular structure elucidation tasks, LLMs are significantly deficient in evaluating solutions as a reward model (see Figure~\ref{fig:intro}(b)).
This is because reward models require an in-depth understanding of the task domain to reflect the complex relationship between predicted molecules and spectral data. LLMs, lacking domain-specific knowledge, are unable to provide precise guidance in assessment and feedback.
}

To address these challenges, we propose a \textbf{K}nowledge-enhanced framework based on Monte Carlo Tree Search (MCTS)
to improve LLMs' reasoning for \textbf{M}olecular \textbf{S}tructure \textbf{E}lucidation (K-MSE).
{To tackle the first issue of limited coverage of chemical molecular space in LLMs, we construct an external molecular substructure knowledge base (Figure~\ref{fig:intro}(a)).
Recognizing molecular substructures serve as the foundational elements of chemical space, this knowledge base integrates both substructures and their structural descriptions through an automated pipeline. Designed to enhance LLMs with domain-specific chemical knowledge, it improves accuracy in inferring molecular structures and reduces errors in atypical cases.
}
{To overcome the second challenge of LLMs inaccurately evaluating solutions, we design a specialized molecule-spectrum scorer as a reward model (Figure~\ref{fig:intro}(b)). This scorer comprises a molecule encoder and a spectrum encoder, which evaluate the alignment between molecular structures and spectral data, providing accurate reward scores for solutions during inference. Furthermore, we integrate this scorer into a Monte Carlo Tree Search (MCTS) reasoning framework combined with Self-Refine~\cite{DBLP:conf/nips/MadaanTGHGW0DPY23}. This framework allows LLMs to optimize previous solutions  in a timely manner while balancing the exploration of new solutions and the exploitation of existing ones.}
By incorporating this scorer into the MCTS framework, LLM can dynamically adjust its reasoning strategy based on feedback, gradually improving the quality of the predicted molecular structure. The scorer also serves as a bridge between the reasoning process and the external knowledge base, using the input spectral data to query and retrieve the most relevant molecular substructures. This effectively mitigates inaccuracies in substructure retrieval, further enhancing the stability and reliability of the model’s reasoning process.

Our contributions can be summarized as follows:
\begin{itemize}
    \item We propose a \textbf{K}nowledge-enhanced reasoning framework for \textbf{M}olecular \textbf{S}tructure \textbf{E}lucidation: K-MSE. It applies test-time scaling {with Monte Carlo Tree Search} and can be integrated with any LLM as a plugin.
    \item We construct an external molecular substructure knowledge base to supplement LLMs' chemical knowledge coverage. Additionally, we design a specialized molecule-spectrum scorer as a reward model to accurately evaluate the reasoning outcomes, while also acting as a retriever between the LLM and the knowledge base.
    \item Experiments on the MolPuzzle benchmark~\cite{guo2024can} demonstrate that our method significantly enhances performance, specifically achieving an improvement of over 20\%
    on GPT-4o-mini and GPT-4o.
\end{itemize}

\section{Related Works}
\paragraph{LLMs for molecular reasoning.}
LLMs have demonstrated significant potential in the field of chemical molecular research~\cite{zhang2025scientific,DBLP:journals/corr/abs-2410-07919}. However, directly applying LLMs to molecular-related tasks presents considerable challenges, primarily due to the inherent complexity of molecules and the substantial differences between molecular structures and natural language~\cite{DBLP:journals/corr/abs-2406-09098,DBLP:conf/acl/JiangZDZC24}. Consequently, many studies have sought to enhance the reasoning of LLMs on molecular data by incorporating domain knowledge. For instance, ChemCrow~\cite{m2024augmenting} leverages existing tools to assist LLMs in various downstream tasks; ChatDrug~\cite{DBLP:conf/iclr/LiuWYW0GX24} integrates domain knowledge into iterative reasoning and feedback loops, enabling LLMs to perform molecular editing tasks; STRUCTCHEM~\cite{DBLP:conf/icml/Ouyang0YL00Q24} uses predefined reasoning templates to guide LLMs through complex chemical tasks. However, to the best of our knowledge, no study has yet explored how to leverage LLMs' reasoning abilities to tackle the challenge of molecular structure elucidation, a crucial problem in chemical experiments that requires deep domain expertise and a precise understanding of molecular structures. The effective application of LLMs to this challenge could significantly enhance the efficiency of scientific experiments and accelerate advancements in chemical research.

\paragraph{LLMs for tree-search reasoning.}
Multi-path reasoning through tree structures, such as Tree-of-Thought (ToT)~\cite{DBLP:conf/nips/YaoYZS00N23} and Monte Carlo Tree Search~\cite{DBLP:conf/emnlp/HaoGMHWWH23,DBLP:conf/emnlp/ChenL0024,DBLP:journals/corr/abs-2406-06592,DBLP:journals/corr/abs-2405-16265}, has proven effective in complex tasks. Some approaches~\cite{DBLP:journals/corr/abs-2408-06195,DBLP:conf/emnlp/HaoGMHWWH23} combine problem decomposition with tree search, where each tree expansion step breaks the problem into manageable subproblems.
However, these methods necessitate LLM's intrinsic capability for problem decomposition and a fine-grained reward model to guide the search process.
In contrast, other studies~\cite{DBLP:journals/corr/abs-2406-07394,DBLP:journals/corr/abs-2410-02884} minimize reliance on problem decomposition by treating the complete solution as a state within the tree, integrating Self-Refine~\cite{DBLP:conf/nips/MadaanTGHGW0DPY23}  to gradually expand the tree structure and optimize reasoning outcomes. Despite their effectiveness, tree search-based reasoning methods highly rely on an accurate reward model to effectively differentiate between desirable and undesirable answers~\cite{DBLP:journals/corr/abs-2211-14275}. Some researchers~\cite{DBLP:journals/corr/abs-2408-06195,DBLP:journals/corr/abs-2406-07394} attempt to leverage the inherent capabilities of LLMs as the reward model, but existing studies have shown that this approach is often limited in its effectiveness~\cite{DBLP:journals/corr/abs-2402-08115}. We have identified that this issue is particularly pronounced in molecular structure elucidation tasks. In this work, we design a specialized scorer as a reward model to provide more accurate guidance signals, thereby enhancing the reasoning process.

\section{Problem Formulation}
Molecular structure elucidation refers to determining a molecule's structure based on various types of spectral data. Following the formulation presented in~\cite{guo2024can}, we define the problem as:
\begin{equation}
\label{eq:def}
    \hat{m}=f(x_\text{ir}, x_\text{cnmr}, x_\text{hnmr}, x_\text{formula}),
\end{equation}
where $\hat{m}$ represents the predicted molecule in SMILES~\cite{weininger1988smiles}, $x_\text{ir}$ is infrared (IR) spectroscopy data, $x_\text{cnmr}$ and $x_\text{hnmr}$ are C-NMR and H-NMR spectra, respectively, and $x_\text{formula}$ denotes molecular formula. The function $f(\cdot)$ represents the reasoning process of LLMs. We denote all input as \textit{question}, where $x_\text{ir}$ is in image modality, while $x_\text{cnmr}$, $x_\text{hnmr}$ and $x_\text{formula}$ are in text modality. We provide an illustrative example in Appendix~\ref{sec:example}.

\section{Method}
\begin{figure}[t]
    \centering
    \includegraphics[width=\linewidth]{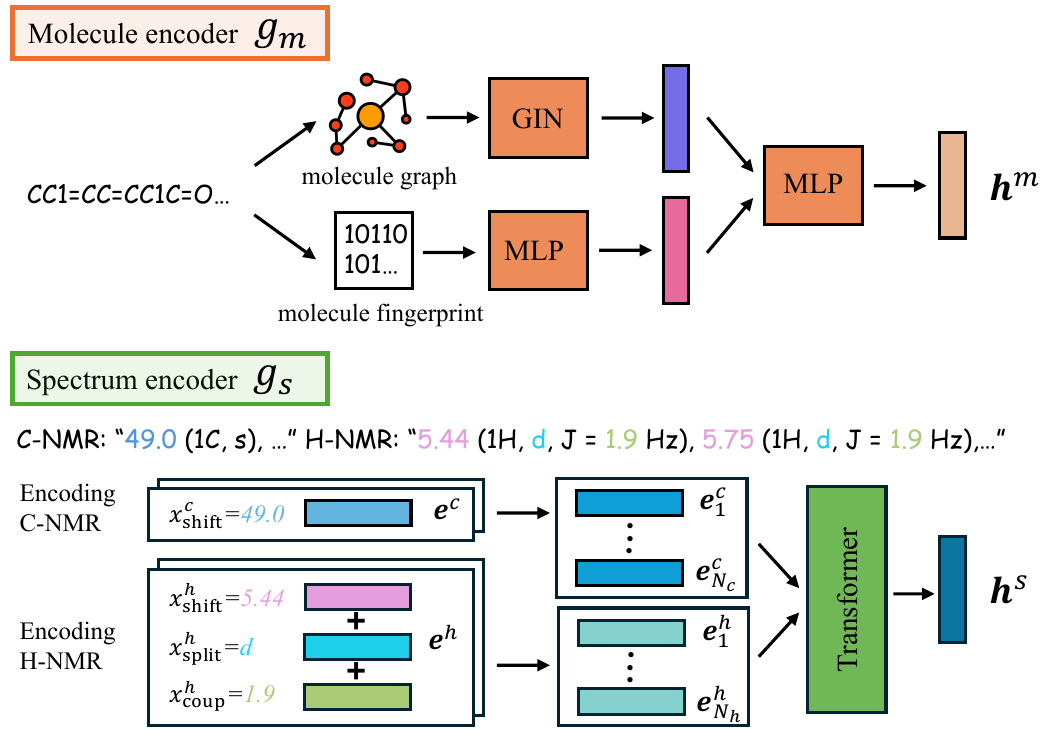} 
    \caption{The molecule-spectrum scorer consists of a molecule encoder $g_m$ and a spectrum encoder $g_s$. After encoding both data types, the reward value is obtained by calculating the similarity between their embeddings $\boldsymbol{h}^m$ and $\boldsymbol{h}^s$.} 
    \label{fig:scorer} 
\end{figure}

\begin{figure*}[t]
    \centering
    \includegraphics[width=0.95\textwidth]{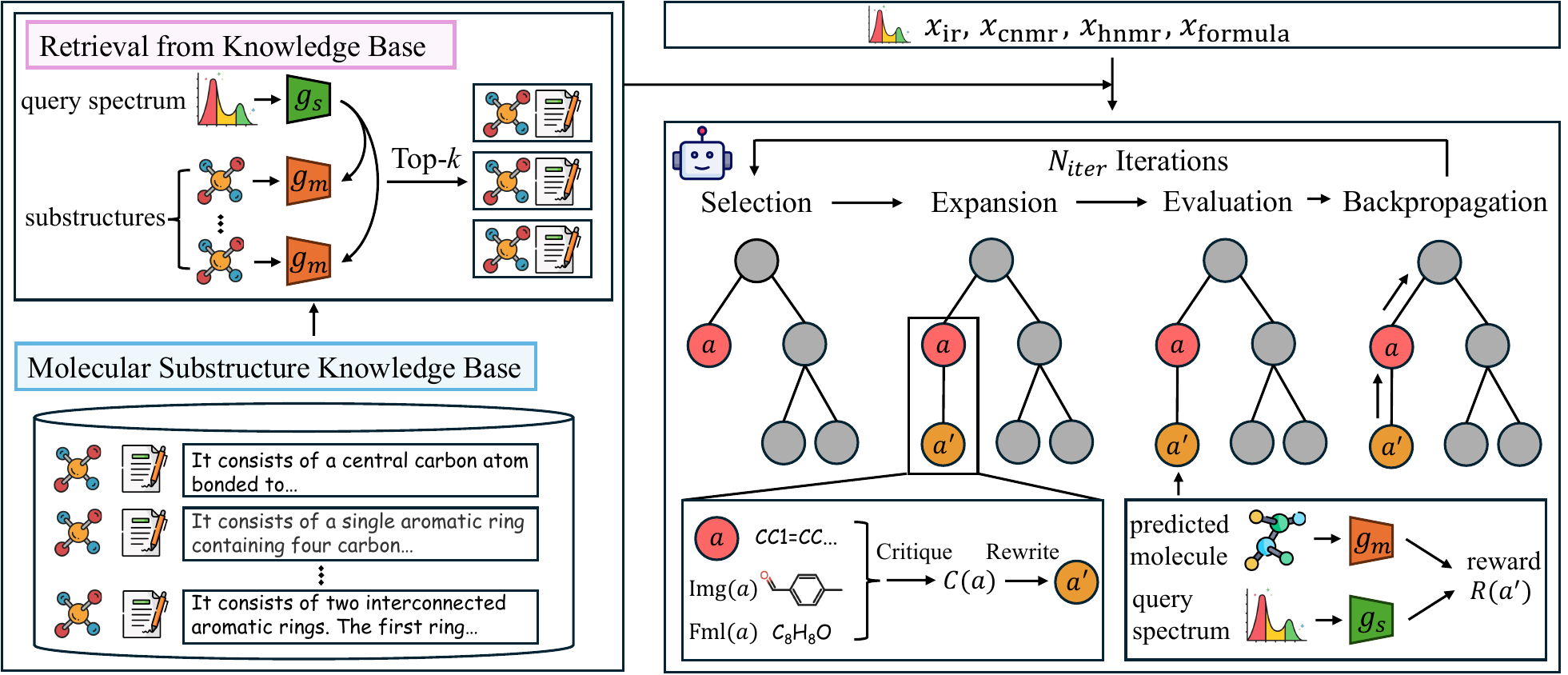} 
    \caption{The framework of K-MSE. We construct a molecular substructure knowledge base covering various substructures and their descriptions. For each \textit{question} including $x_\text{ir}$, $x_\text{cnmr}$, $x_\text{hnmr}$, $x_\text{formula}$, relevant information is first retrieved from the knowledge base. Then, an MCTS-based iteration is performed, where each node in the tree represents a complete answer. The expansion process critiques previous answers $a$ with image $\mathrm{Img}(a)$ and formula $\mathrm{Fml}(a)$ of the predicted molecule, followed by a rewrite.  We derive the reward of answers using the specialized scorer containing $g_s$ and $g_m$, which is also applied during the retrieval process.} 
    \label{fig:framework} 
\end{figure*}

In this section, we present the proposed \textbf{K}nowledge-enhanced reasoning framework for \textbf{M}olecular \textbf{S}tructure \textbf{E}lucidation (K-MSE). First, we describe the constructed knowledge base, which contains molecular substructures to expand the chemical structure coverage of LLMs. Next, we introduce the molecule-spectrum scorer (Figure~\ref{fig:scorer}), a specialized reward model that provides accurate guidance during reasoning and serves as a bridge to retrieve from the knowledge base. Finally, we outline the overall reasoning framework (Figure~\ref{fig:framework}).

\subsection{Molecular Substructure Knowledge Base}

The molecular structure space is vast and complex, estimated to contain approximately $10^{60}$ molecules~\cite{polishchuk2013estimation}. This immense space poses a significant challenge for LLMs in molecular structure prediction, primarily due to the high diversity of molecular structures and the inherent differences between chemical structures and natural language representations~\cite{DBLP:conf/acl/JiangZDZC24}. LLMs are often adept at recognizing basic molecular structures but struggle to make accurate predictions when faced with more complex situations (exhibited in Figure~\ref{fig:intro}(a)). In other words, LLMs exhibit limitations in achieving a comprehensive coverage of molecular structures.

To address this issue, we construct a molecular substructure knowledge base. Substructures refer to common fragments in molecules that exhibit specific functional or characteristic features, serving as fundamental units for molecular analysis~\cite{fang2023knowledge}. This knowledge base contains a broad array of molecular substructures along with their corresponding textual descriptions, which can dynamically provide valuable supplementary information to the LLM during the reasoning process. By leveraging this knowledge base, we equip the LLM with rich, domain-specific chemical knowledge, thereby enhancing its ability to more accurately identify and infer complex molecular structures.

Specifically, the knowledge base $\mathcal{KB}$ can be formulated as:
\begin{equation}
    \mathcal{KB}=\{(s_i,d_i)\},
\end{equation}
where $s_i$ represents the molecular substructure in SMILES~\cite{weininger1988smiles}, and $d_i$ is the corresponding textual description.
{In $\mathcal{KB}$, both ring and chain substructures are extracted from a widely used molecule database~\cite{polykovskiy2020molecular}, achieving a balance between diversity and universality. By incorporating structural information obtained through external tools as auxiliary input, we use LLM to automate the generation of reliable descriptions for these substructures.}
The construction details are in Appendix~\ref{sec:appendix-kb} and Figure~\ref{fig:kb}.

\subsection{Molecule-Spectrum Scorer}
A reliable reward model is crucial for ensuring that LLMs can effectively correct errors and improve their reasoning capabilities~\cite{DBLP:journals/corr/abs-2211-14275}. The reward model provides precise feedback signals, guiding LLMs to adjust suboptimal reasoning paths, thus progressively refining the final result. However, as revealed in Figure~\ref{fig:intro}(b), leveraging the inherent capabilities of LLMs as a reward model for molecular structure elucidation tasks has limitations. This is primarily due to the lack of deep chemical knowledge in LLMs, which prevents them from accurately assessing the consistency between predicted molecular structures and queried spectral data, making it challenging to differentiate between ideal and non-ideal solutions.

To provide accurate guidance for reasoning, we design a specialized scorer as the reward model. This model consists of a molecule encoder $g_{m}$ and a spectrum encoder $g_{s}$:
\begin{equation}
    \boldsymbol{h}^{m}=g_{m}(m),\; \boldsymbol{h}^{s}=g_{s}(n),
\end{equation}
where $m$ is molecule, $n$ is NMR spectrum $x_\text{cnmr}$ and $x_\text{hnmr}$ in Equation~\eqref{eq:def}. $\boldsymbol{h}^{m},\boldsymbol{h}^{s}\in \mathbb{R}^d$ are embeddings for molecule and spectrum, respectively.

Typically, the molecule encoder $g_{m}$ takes both molecule graph $m_{g}$~\cite{DBLP:conf/iclr/HuLGZLPL20} and fingerprint $m_{fp}$~\cite{rogers2010extended} as input, and employs GIN~\cite{DBLP:conf/iclr/XuHLJ19} and MLP for encoding:
\begin{equation}
    \boldsymbol{h}^{m}=\mathrm{MLP}(\mathrm{GIN}(m_{g}), \mathrm{MLP}(m_{fp})).
\end{equation}

For spectrum, we design a model that leverages the chemical shift of each carbon atom in C-NMR and the chemical shift, splitting pattern, and coupling constants of each hydrogen atom in H-NMR.
For example, for a carbon atom with C-NMR spectrum \textit{``49.0 (1C, s)''}, its chemical shift is \textit{49.0}. For a hydrogen atom with H-NMR spectrum \textit{``5.44 (1H, d, J = 1.9 Hz)''}, its chemical shift, splitting pattern, and coupling constant are \textit{5.44}, \textit{d}, and \textit{1.9}, respectively.
To represent these features, we discretize all possible values for chemical shifts and coupling constants, assigning a unique token ID to each value. Similarly, we assign a separate token ID to each splitting pattern. The formal representation is as follows:
\begin{equation}
\begin{aligned}
    \boldsymbol{e}^c &= \mathrm{E}(x^c_{\text{shift}}), \\
    \boldsymbol{e}^h &= \mathrm{E}(x^h_{\text{shift}}) + \mathrm{E}(x^h_{\text{split}}) + \mathrm{E}(x^h_{\text{coup}}),
\end{aligned}
\end{equation}
where $\boldsymbol{e}^c$ and $\boldsymbol{e}^h$ are embeddings for one carbon and hydrogen atom, respectively. $x^c_{\text{shift}}$ is the carbon chemical shift, $x^h_{\text{shift}}$, $x^h_{\text{split}}$ and $x^h_{\text{coup}}$ are hydrogen chemical shift, splitting pattern and coupling constant. $\mathrm{E}(\cdot)$ denotes the embedding function. We then concatenate the embeddings of all carbon and hydrogen atoms and input them into a Transformer encoder~\cite{DBLP:conf/nips/VaswaniSPUJGKP17}:
\begin{equation}
    \boldsymbol{h}^{s}=\mathrm{Transformer}(\boldsymbol{e}^c_1,...,\boldsymbol{e}^c_{N_{c}},\boldsymbol{e}^h_1,...,\boldsymbol{e}^h_{N_{h}}),
\end{equation}
where $N_{c}$ and $N_{h}$ are numbers of carbon and hydrogen atoms in this molecule.

For model training, the NT-Xent loss~\cite{DBLP:conf/icml/ChenK0H20} is employed:
\begin{equation}
        \mathcal{L} = -\sum\limits_{i=1}^B \frac{e^{\mathrm{sim}(\boldsymbol{h}^{m}_i,\boldsymbol{h}^{s}_i)/\tau}}{  \sum\limits_{k=1,k\ne i}^{B} e^{\mathrm{sim}(\boldsymbol{h}^{m}_i,\boldsymbol{h}^{s}_k)/\tau} + e^{\mathrm{sim}(\boldsymbol{h}^{m}_k,\boldsymbol{h}^{s}_i)/\tau}},
\end{equation}
where $\boldsymbol{h}^{m}_i$ and $\boldsymbol{h}^{s}_i$ are molecule and spectrum embeddings for the $i$-th molecule. $\mathrm{sim(\cdot,\cdot)}$ denotes cosine similarity, $B$ is the batch size, and $\tau$ is temperature. Details of the model architecture and training process are in Appendix~\ref{sec:appendix-scorer}.

\subsection{MCTS-based Reasoning Framework}
{Effective reasoning requires both knowledge mastery and timely reflection and adjustment. We expand knowledge coverage via a knowledge base and introduce a scorer to evaluate reasoning outcomes accurately. However, LLMs lack self-reflection and dynamic reasoning adjustment, hindering proactive error detection and correction during generation.
To address this limitation, we integrate the knowledge base and scorer into a MCTS reasoning framework, which treats each inference process and solution as a tree node~\cite{DBLP:journals/corr/abs-2410-02884}, and employs Self-Refine~\cite{DBLP:conf/nips/MadaanTGHGW0DPY23} for iterative reflection and improvement during tree search. Prior to MCTS, we first retrieve from the knowledge base. Then MCTS operates iteratively, with each iteration consisting of four key steps: selection, expansion, evaluation, and backpropagation.
}

\paragraph{Retrieval from Knowledge Base.}
First, based on the spectrum data in the \textit{question}, we retrieve the most relevant substructures and their corresponding descriptions from the molecular substructure knowledge base. Since the objective of the proposed molecule-spectrum scorer is to measure the correlation between spectral data and molecular structures, it effectively acts as a bridge connecting the query and the knowledge base.  Specifically, we use the spectrum encoder $g_{s}$ to encode the query spectrum and use the molecule encoder $g_{m}$ to encode each substructure in the knowledge base. Retrieval is employed as follows:
\begin{equation}
    \{(s_j,d_j)\} = \underset{s_i\in\mathcal{KB}}{\mathrm{Top}\text{-}k}(\mathrm{sim}(g_{m}(s_i),
    g_{s}(n))),
\end{equation}
where $n$ is spectrum $x_\text{cnmr}$ and $x_\text{hnmr}$ in Equation~\eqref{eq:def}. The retrieved $\{(s_j,d_j)\}$ are then appended after the \textit{question} for subsequent reasoning.

\paragraph{Selection.}
In the iteration of MCTS, the first step is to select a node for expansion. We use the Upper Confidence Bound applied to Trees~(UCT) as selection criterion:
\begin{equation}
\label{eq:uct}
    a=\arg \max (Q(a_i)+c\sqrt{\frac{\ln{N(P(a_i))} +1}{N(a_i)+\varepsilon } } ),
\end{equation}
where $Q(a_i)\in \mathbb{R}$ is the Q-value of node $a_i$, which is initialized in the evaluation step and updated in the backpropagation step; $N(\cdot)$ is the number of visits; and $P(\cdot)$ is the parent node. $c$ and $\varepsilon$ are hyper-parameters. A node is considered fully selected when its child node number reaches a predefined limit.

\paragraph{Expansion.}
In expansion, a critique process is employed to identify the deficiencies of the current solution in the selected node $a$:
\begin{equation}
\label{eq:critique}
    C(a)=\mathrm{Critique}(a, \mathrm{Img}(a), \mathrm{Fml}(a)),
\end{equation}
where $\mathrm{Img}(a)$ and $\mathrm{Fml}(a)$ represent image and chemical formula of the predicted molecule in $a$. We observe that for LLMs, the effectiveness of critiquing previous response without providing additional context is significantly limited, as they often struggle to accurately identify molecular structures from text alone. This issue can be mitigated by supplementing with molecular image and chemical formula. Subsequently, a rewrite process is employed to produce a new solution $a^{\prime}$ as the child of $a$, based on $C(a)$:
\begin{equation}
    a^{\prime}=\mathrm{Rewrite}(a, C(a)).
\end{equation}

\paragraph{Evaluation.}
The newly generated node necessitates an evaluation as its reward value. We employ the proposed molecule-spectrum scorer as the reward model to compute the similarity between the predicted molecular structure and the queried spectrum:
\begin{equation}
    R(a^\prime)=\mathrm{sim}(g_m(m_{a^\prime}),g_s(n)),
\end{equation}
where $m_{a^\prime}$ is the predicted molecule in $a^\prime$, $n$ represents spectrum $x_\text{cnmr}$ and $x_\text{hnmr}$, and $R(a^\prime) \in \mathbb{R}$ is the reward value. The Q-value of $a^\prime$ is then  initialized with the reward: $Q(a^\prime)=R(a^\prime)$.

\paragraph{Backpropagation.}
In the backpropagation step, the Q-value of the new node is propagated to its parent node:
\begin{equation}
    Q(a)=0.5\times Q(a^\prime)+0.5\times Q(a).
\end{equation}
This backpropagation process is iteratively applied until reaching the root. Backpropagation enables the adjustment of node value estimates based on newly acquired feedback, progressively refining the reasoning process and guiding it toward more promising regions of the solution space.

The number of iterations for selection, expansion, evaluation, and backpropagation is controlled by the hyper-parameter $N_{iter}$. Upon reaching the maximum number of iterations, the node with the highest reward is selected as the final answer.

\section{Experiment}

\subsection{Experimental Setup}
We conduct experiments on the MolPuzzle benchmark~\cite{guo2024can}, following the ``Addressing Entire Molecule Puzzles'' setting. IR, C-NMR, H-NMR, and molecular formula serve as inputs, and the goal is to predict the molecule in a \textit{zero-shot} manner. The dataset consists of 216 molecules. We select the following base models: Llama-3.2-11B-Vision-Instruct~\cite{dubey2024llama}, GPT-4o-mini (\texttt{gpt-4o-mini-2024-07-18})~\cite{openai4omini}, GPT-4o (\texttt{gpt-4o-2024-08-06})~\cite{openai4o}, and GPT-o1 (\texttt{o1-2024-12-17})~\cite{openaio1}.

The baseline methods include direct prompting with Chain-of-Thought (CoT)~\cite{DBLP:conf/nips/Wei0SBIXCLZ22}, as well as employing Self-Refine~(Self-R)~\cite{DBLP:conf/nips/MadaanTGHGW0DPY23}, Self-Consistency~(Self-C)~\cite{DBLP:conf/iclr/0002WSLCNCZ23}, Multi-Agent Debate~(MAD)~\cite{DBLP:conf/emnlp/Liang0JW00Y0T24}, and MCT Self-Refine~(MCTSr)~\cite{DBLP:journals/corr/abs-2406-07394}. Evaluation metrics include: (1) Fingerprint Tanimoto Similarity (FTS)~\cite{tanimoto1958elementary} for chemical similarity between generated and ground-truth molecules using Morgan, MACCS, and RDK fingerprints; (2) Formula accuracy (Formula Acc), which measures the correctness of chemical formulas of generated molecules; and (3) ACC for the exact structural match. Detailed implementation of the proposed K-MSE and baselines are in Appendix~\ref{sec:appendix-impl} and \ref{sec:appendix-baseline}.

\begin{table*}[t]
\centering
\scalebox{0.9}{
\begin{tabular}{lccccc}
\toprule
Model & Morgan FTS & MACCS FTS & RDK FTS & Formula ACC & ACC \\ \midrule
Llama-3.2-11B & 0.163 & 0.349 & 0.188 & 0.060 & 0.014 \\
+ Self-R & 0.161 \scalebox{0.9}{($-$0.002)} & 0.345 \scalebox{0.9}{($-$0.004)} & 0.178 \scalebox{0.9}{($-$0.010)} & 0.023 \scalebox{0.9}{($-$0.037)} & 0.023 \scalebox{0.9}{($+$0.009)} \\
+ Self-C & 0.172 \scalebox{0.9}{($+$0.009)} & 0.350 \scalebox{0.9}{($+$0.001)} & 0.210 \scalebox{0.9}{($+$0.022)} & 0.032 \scalebox{0.9}{($-$0.028)} & 0.019 \scalebox{0.9}{($+$0.005)} \\
+ MAD & 0.185 \scalebox{0.9}{($+$0.022)} & 0.368 \scalebox{0.9}{($+$0.019)} & 0.203 \scalebox{0.9}{($+$0.015)} & 0.051 \scalebox{0.9}{($-$0.009)} & 0.009 \scalebox{0.9}{($-$0.005)} \\
+ MCTSr & 0.169 \scalebox{0.9}{($+$0.006)} & 0.343 \scalebox{0.9}{($-$0.006)} & 0.197 \scalebox{0.9}{($+$0.009)} & 0.023 \scalebox{0.9}{($-$0.037)} & 0.004 \scalebox{0.9}{($-$0.010)} \\
+ K-MSE & \textbf{0.298 \scalebox{0.9}{($+$0.135)}} & \textbf{0.465 \scalebox{0.9}{($+$0.116)}} & \textbf{0.311 \scalebox{0.9}{($+$0.123)}} & \textbf{0.143 \scalebox{0.9}{($+$0.083)}} & \textbf{0.111 \scalebox{0.9}{($+$0.097)}} \\ \midrule
GPT-4o-mini & 0.260 & 0.512 & 0.337 & 0.185 & 0.037 \\
+ Self-R & 0.287 \scalebox{0.9}{($+$0.027)} & 0.523 \scalebox{0.9}{($+$0.011)} & 0.369 \scalebox{0.9}{($+$0.032)} & 0.231 \scalebox{0.9}{($+$0.046)} & 0.069 \scalebox{0.9}{($+$0.032)} \\
+ Self-C & 0.299 \scalebox{0.9}{($+$0.039)} & 0.535 \scalebox{0.9}{($+$0.023)} & 0.373 \scalebox{0.9}{($+$0.036)} & 0.222 \scalebox{0.9}{($+$0.037)} & 0.083 \scalebox{0.9}{($+$0.046)} \\
+ MAD & 0.292 \scalebox{0.9}{($+$0.032)} & 0.520 \scalebox{0.9}{($+$0.008)} & 0.359 \scalebox{0.9}{($+$0.022)} & 0.222 \scalebox{0.9}{($+$0.037)} & 0.079 \scalebox{0.9}{($+$0.042)} \\
+ MCTSr & 0.281 \scalebox{0.9}{($+$0.021)} & 0.530 \scalebox{0.9}{($+$0.018)} & 0.361 \scalebox{0.9}{($+$0.024)} & 0.176 \scalebox{0.9}{($-$0.009)} & 0.069 \scalebox{0.9}{($+$0.032)} \\
+ K-MSE & \textbf{0.470 \scalebox{0.9}{($+$0.210)}} & \textbf{0.651 \scalebox{0.9}{($+$0.139)}} & \textbf{0.520 \scalebox{0.9}{($+$0.183)}} & \textbf{0.412 \scalebox{0.9}{($+$0.227)}} & \textbf{0.273 \scalebox{0.9}{($+$0.236)}} \\ \midrule
GPT-4o & 0.493 & 0.690 & 0.538 & 0.486 & 0.278 \\
+ Self-R & 0.474 \scalebox{0.9}{($-$0.019)} & 0.692 \scalebox{0.9}{($+$0.002)} & 0.522 \scalebox{0.9}{($-$0.016)} & 0.500 \scalebox{0.9}{($+$0.014)} & 0.250 \scalebox{0.9}{($-$0.028)} \\
+ Self-C & 0.551 \scalebox{0.9}{($+$0.058)} & 0.732 \scalebox{0.9}{($+$0.042)} & 0.581 \scalebox{0.9}{($+$0.043)} & 0.514 \scalebox{0.9}{($+$0.028)} & 0.347 \scalebox{0.9}{($+$0.069)} \\
+ MAD & 0.519 \scalebox{0.9}{($+$0.026)} & 0.710 \scalebox{0.9}{($+$0.020)} & 0.558 \scalebox{0.9}{($+$0.020)} & 0.482 \scalebox{0.9}{($-$0.004)} & 0.310 \scalebox{0.9}{($+$0.032)} \\
+ MCTSr & 0.500 \scalebox{0.9}{($+$0.007)} & 0.705 \scalebox{0.9}{($+$0.015)} & 0.551 \scalebox{0.9}{($+$0.013)} & 0.495 \scalebox{0.9}{($+$0.009)} & 0.282 \scalebox{0.9}{($+$0.004)} \\
+ K-MSE & \textbf{0.707 \scalebox{0.9}{($+$0.214)}} & \textbf{0.834 \scalebox{0.9}{($+$0.144)}} & \textbf{0.727 \scalebox{0.9}{($+$0.189)}} & \textbf{0.711 \scalebox{0.9}{($+$0.225)}} & \textbf{0.578 \scalebox{0.9}{($+$0.300)}} \\ \midrule
GPT-o1 & 0.833 & 0.923 & 0.856 & 0.903 & 0.741 \\
+ K-MSE & \textbf{0.868 \scalebox{0.9}{($+$0.035)}} & \textbf{0.939 \scalebox{0.9}{($+$0.016)}} & \textbf{0.888 \scalebox{0.9}{($+$0.032)}} & \textbf{0.923 \scalebox{0.9}{($+$0.020)}} & \textbf{0.787 \scalebox{0.9}{($+$0.046)}} \\ \bottomrule
\end{tabular}
}
\caption{Experimental performance on MolPuzzle benchmark. The first row of each group presents the baseline results by directly employing CoT, with the values in $()$ indicating the relative improvement over the baseline. The best results are highlighted in \textbf{bold}.}
\label{tab:main-result}
\end{table*}

\subsection{Main Results}
Experimental results are presented in Table~\ref{tab:main-result}.
Our method, K-MSE, significantly outperforms all baseline methods, achieving substantial performance improvements across all base models. Specifically, ACC increases by 0.236 and 0.300 on GPT-4o-mini and GPT-4o, respectively. As the capability of the base models improves, the relative performance gain of K-MSE exhibits an initial increase followed by a decline.  We attribute this to that weaker base models (e.g., Llama-3.2-11B) have limited capabilities in chemical reasoning, preventing them from fully exploiting the potential of K-MSE. Conversely, stronger models (e.g., GPT-o1) are closer to the performance ceiling, resulting in diminishing returns with further optimization.

Additionally, some baseline methods lead to performance degradation. For example, MAD and MCTSr cause ACC to decrease by 0.005 and 0.010, respectively, on Llama-3.2-11B. This can be attributed to the insufficient knowledge and reasoning abilities of LLMs in chemical tasks. Both MAD and MCTSr rely on the model’s self-correction mechanism, which requires accurate understanding and adjustment of the reasoning process. Over-reliance on this potentially unreliable mechanism can amplify errors, resulting in greater discrepancies between the generated and ground-truth molecular structures.
In contrast, our approach enhances domain knowledge by incorporating a molecular substructure knowledge base, utilizes specialized scorer to evaluate reasoning outcomes, and introduces additional inputs (such as molecular image and formula) during the self-critique phase. These ensure that K-MSE achieves consistent and significant performance gains across all base models. Further discussion on these can be found in subsequent Sections and Appendix~\ref{sec:appendix-results}.

\subsection{Detailed Analysis}

\begin{figure}[t]
    \centering
    \includegraphics[width=\linewidth]{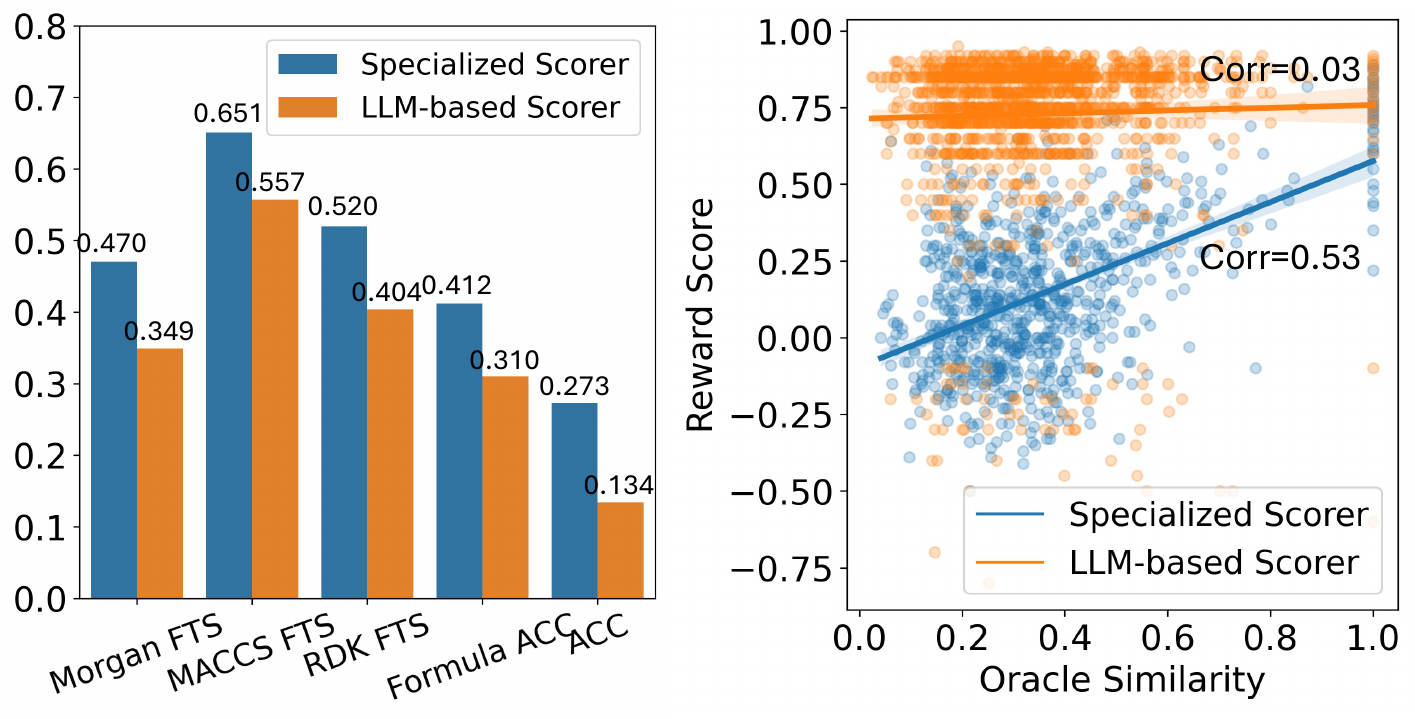} 
    \caption{Performance comparison (left) and reward vs. oracle similarity (right) between specialized scorer and LLM-based scorer on GPT-4o-mini.} 
    \label{fig:ablation-scorer} 
\end{figure}

\paragraph{Specialized Scorer Ensures Precise Answer Evaluation.}
To evaluate the specialized scorer, we compare it with an LLM-based scorer. The experimental results, shown in Figure~\ref{fig:ablation-scorer}, highlight the performance differences between the two approaches and analyze the distribution and correlation between the reward score and oracle similarity. The reward score is the score assigned by the scorer to each prediction, while the oracle represents the molecular fingerprint similarity between the predicted and the ground-truth molecules. An ideal scorer should exhibit a positive correlation between these two measures.
The results clearly show that the specialized scorer significantly enhances performance, yielding an ACC increase of 0.139. Notably, the specialized scorer shows a positive correlation of 0.53 between the reward score and the oracle.
However, the LLM-based scorer shows a near-zero correlation (0.03) and produces less discriminative, overly lenient scores (generally >0.7), likely due to LLM self-affirmation bias. 
In contrast, the specialized scorer, as a tailored small model, effectively addresses this bias, providing more accurate and reliable evaluations.

\begin{figure}[t]
    \centering
    \includegraphics[width=\linewidth]{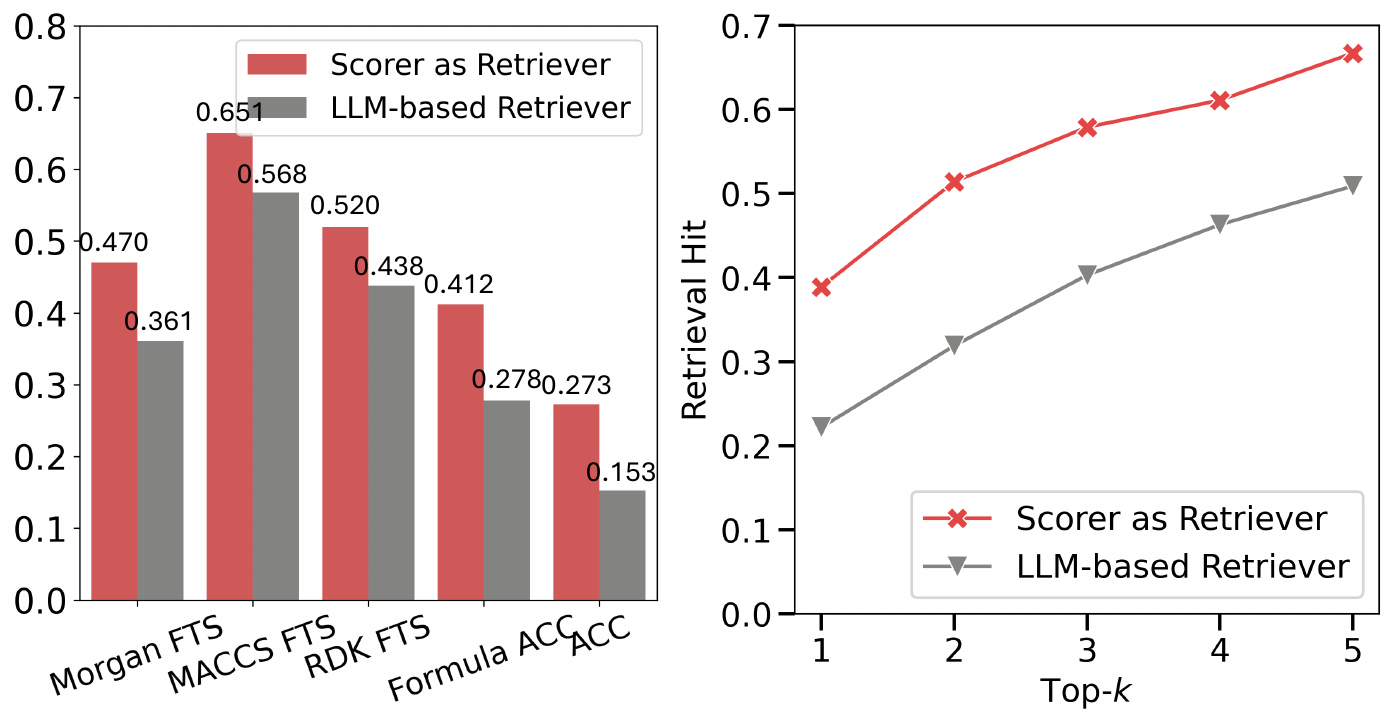} 
    \caption{Performance comparison (left) and retrieval hit comparison (right) between scorer as retriever and LLM-based retriever on GPT-4o-mini.} 
    \label{fig:ablation-retriever} 
\end{figure}

\paragraph{Specialized Scorer Enhances Knowledge Base Retrieval.}
In K-MSE, the specialized scorer also acts as the retriever between LLM and knowledge base. To assess its effectiveness, we
conduct experiments with an LLM-based retriever variant. In this variant, we first prompt the LLM to analyze the molecular structure from the \textit{question} and generate a query description. Then, we use BM25~\cite{robertson2009probabilistic} to compute the similarity between the query description and the substructure descriptions in the knowledge base.
Experimental results in Figure~\ref{fig:ablation-retriever} demonstrate that the scorer-based retriever significantly outperforms the LLM-based approach, with a 0.120 improvement in ACC and an average 0.169 increase in retrieval hit rate (the proportion of retrieved substructures containing the ground-truth substructures in the target molecule). This highlights the scorer's ability to retrieve more accurate substructures with fewer retrievals.
\begin{figure}[t]
    \centering
    \includegraphics[width=\linewidth]{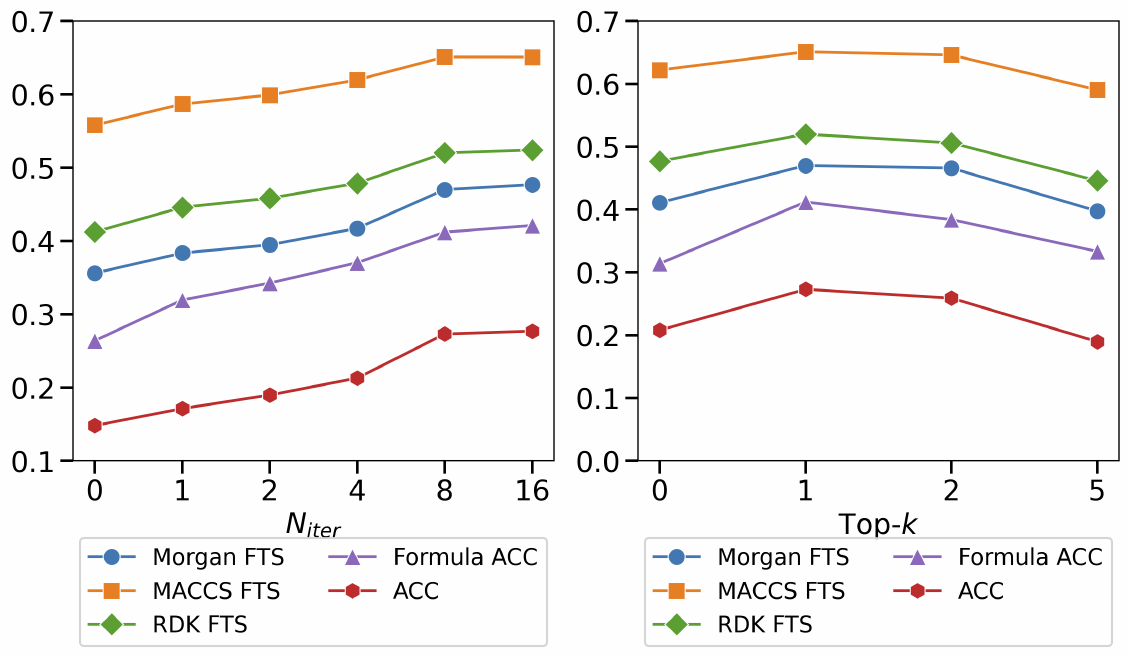} 
    \caption{Analysis of MCTS iterations $N_{iter}$ (left) and retrieval counts $k$ from the knowledge base (right).} 
    \label{fig:ablation-iter-topk} 
\end{figure}

\paragraph{MCTS Iterations and Knowledge Base Retrieval Analysis.}
We analyze the impact of different iteration numbers $N_{iter}$ and retrieval counts $k$ on performance using GPT-4o-mini. Experimental results are shown in Figure~\ref{fig:ablation-iter-topk}.
Setting $N_{iter}$ being 0 disables the MCTS iterative process, while $k$ being 0 removes the knowledge base. 
Results show that performance improves with increasing $N_{iter}$, demonstrating the benefits of iterative reasoning, though gains diminish after a certain point.
For retrieval count $k$, performance initially rises but then declines as $k$ grows, suggesting that while the knowledge base integration is effective, larger $k$ values introduce noise from retrieved irrelevant substructures. 
Notably, the best performance on GPT-4o-mini occurs at $k=1$, while on GPT-4o, optimal performance is achieved at $k=2$. This difference may stem from GPT-4o's greater robustness, which allows it to more effectively mitigate the impact of retrieval noise.

\section{Conclusion}
In this work, we introduce K-MSE, a knowledge-enhanced reasoning framework for molecular structure elucidation based on LLMs. K-MSE utilizes a Monte Carlo Tree Search-based test-time scaling approach and can be integrated as a plugin with any LLM. We build a molecular substructure knowledge base and design a specialized molecule-spectrum scorer to address the limited chemical knowledge inherent in LLMs. Experimental results demonstrate the superior performance of K-MSE, and we believe it paves the way for LLMs to function as copilots in the analysis of chemical experiments, thereby accelerating scientific advances.

\clearpage
\section*{Limitations}
Although the proposed K-MSE framework demonstrates significant performance improvements in molecular structure elucidation, several limitations should be acknowledged. First, the framework operates as a plugin for existing large language models (LLMs) rather than training LLMs specifically for the task. While this design choice enhances flexibility and reduces dependency on task-specific training, the performance of K-MSE is inherently constrained by the capabilities of the underlying LLM. This limitation is particularly pronounced due to the scarcity of reliable and publicly available molecular spectral data, which hinders the development of more specialized models. Second, the evaluation of K-MSE is currently limited to the MolPuzzle benchmark, as it is the only publicly accessible dataset through reliable experimental studies in this domain. While MolPuzzle provides a valuable starting point, its limited scale and diversity may not fully capture the complexity of real-world molecular structure elucidation tasks. To address these limitations, future work will focus on (1) collecting a larger and more diverse dataset to enable more robust training and evaluation, and (2) incorporating cutting-edge LLM training strategies, such as slow-thinking approaches, to further enhance LLM's reasoning capabilities.

\section*{Acknowledgements}
This work is funded by the “Pioneer” and “Leading Goose”
R\&D Program of Zhejiang (Grant No. 2025C01097), NSFCU23B2055, NSFC2302433, NSFCU23A20496, the Fundamental Research Funds for the Central Universities (226-2023-00138), Zhejiang Provincial Natural Science Foundation of China (LQ24F020007) and Hangzhou West Lake Pearl Project Leading Innovative Youth Team Project (TD2023017).
\bibliography{main}

\clearpage

\appendix

\begin{figure*}[!t]
    \centering
    \includegraphics[width=\textwidth]{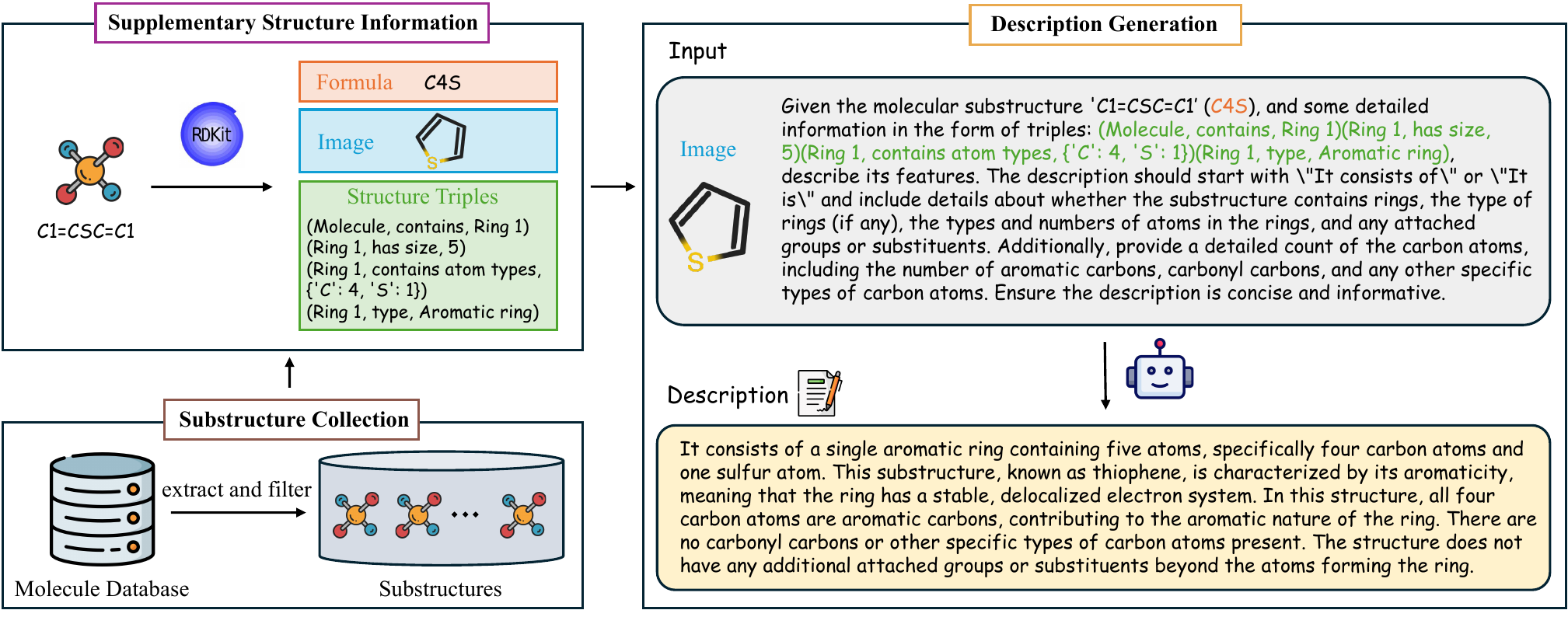} 
    \caption{Construction process of the molecular substructure knowledge base. The process consists of three steps: First, collect molecular substructures from a molecule database. Then, RDKit is used to obtain supplementary structure information for each substructure, including molecular formula, structure image, and structure triplets. Finally, LLM is prompted to generate the corresponding textual descriptions.} 
    \label{fig:kb} 
\end{figure*}

\section{Implementation of K-MSE}

\subsection{Construction of Molecular Substructure Knowledge Base}
\label{sec:appendix-kb}
The molecular substructures in the knowledge base are divided into two parts: ring structures and chain structures. A ring structure refers to a closed loop of atoms, such as a benzene ring (C1=CC=CC=C1), where the atoms form a cyclic arrangement. A chain structure, on the other hand, refers to a linear arrangement of atoms within the molecule, such as a propyl group (CCC), where the atoms are connected in a straight or branched chain. The extraction of ring structures is defined as follows: first, individual ring structures within the molecule are identified. Then, rings that share two or more atoms are merged. Finally, atoms that are connected to the ring atoms via chemical bonds are included. The extraction of chain structures is defined as follows: first, carbon atoms that are not part of any ring are identified. Then, for each carbon atom, the substructures within its two-hop neighbors are extracted. Finally, substructures containing atoms that are part of a ring are filtered out. We collect substructures from over 4 million molecules in the Moses database~\cite{polykovskiy2020molecular} using the extraction methods described above. Substructures that appear fewer than 1,000 times are filtered out to maintain a balance between diversity and universality. Ultimately, the knowledge base contains 593 substructures, and substructures are represented in SMILES format.

\begin{figure}[t]
    \centering
    \includegraphics[width=\linewidth]{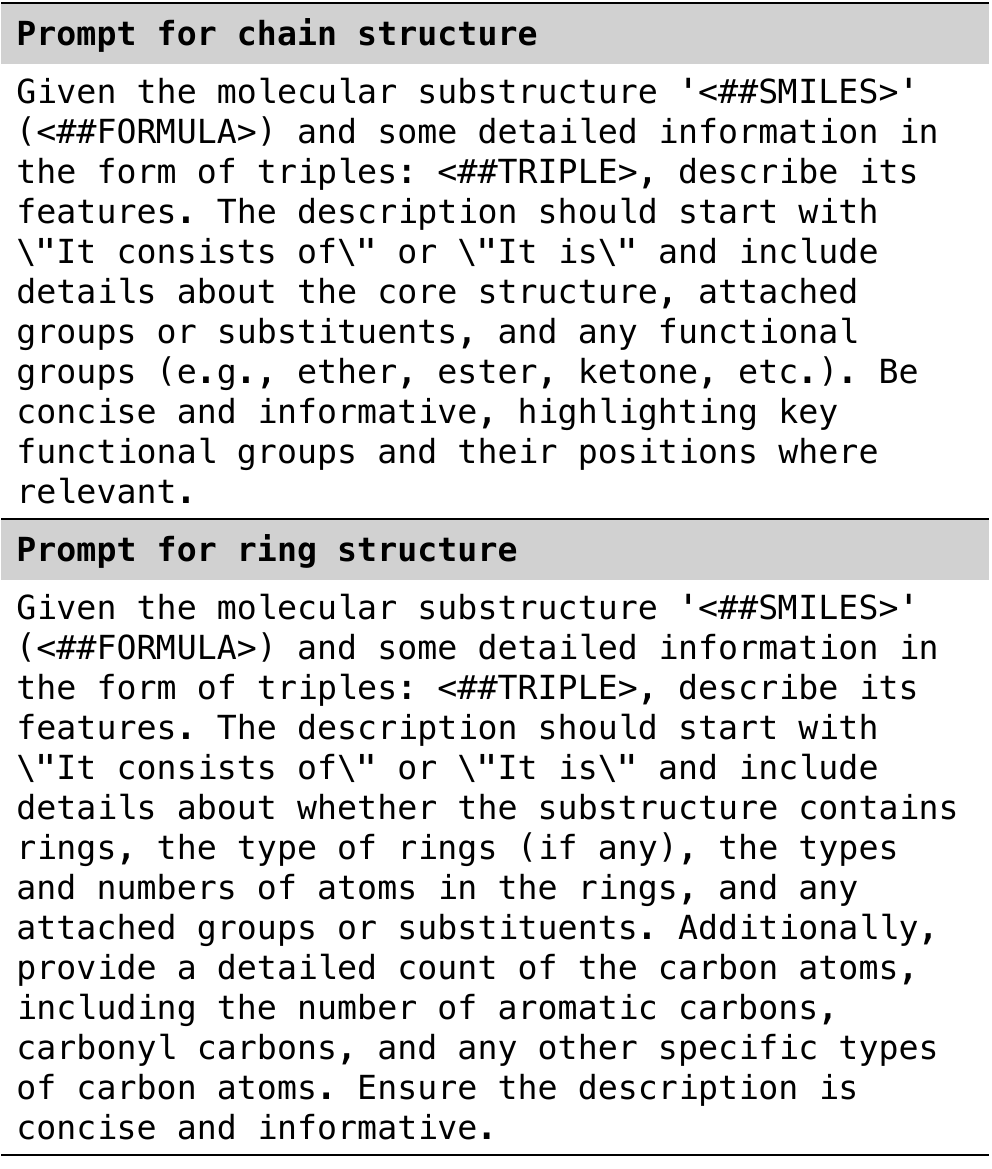} 
    \caption{Prompt for generating descriptions in construction of molecular substructure knowledge base.} 
    \label{fig:prompt-db} 
\end{figure}

Another crucial component of the knowledge base is the natural language description of molecular substructures. We use LLMs to automate the process of describing these substructures. Since LLMs may not always accurately recognize molecular structures from SMILES notation, we complement the input by incorporating structural information obtained through external tools. These supplementary structural details include triplet-form structure descriptions, molecular formula, and molecular image.
For the triplet-form structure descriptions, we leverage RDKit to extract molecular structure information using the SMILES notation as input. This information includes the number of rings, the size and type of each ring, the types of atoms in the rings, and the functional groups present, such as ether, ester, ketone, imine, and nitrile, etc. Additionally, the corresponding molecular formula and image are also obtained using RDKit.
In our implementation, we use GPT-4o-mini to generate natural language descriptions of molecular structures. Our manual spot checks have shown that LLM can provide accurate structural descriptions by incorporating external tools in this manner. The entire knowledge base construction process is presented in Figure~\ref{fig:kb}, and the prompt used in description generation is in Figure~\ref{fig:prompt-db}.

\subsection{Details of Specialized Molecule-Spectrum Scorer}
\label{sec:appendix-scorer}

\paragraph{Model architecture.}
The scorer consists of a molecule encoder $g_m$ and a spectrum encoder $g_s$, which are responsible for encoding the molecular and spectral data, respectively. The molecule encoder $g_m$ includes a GIN (Graph Isomorphism Network)~\cite{DBLP:conf/iclr/XuHLJ19} for encoding the molecular graph and an MLP (Multilayer Perceptron) for encoding the molecular fingerprint. The molecular graph can be obtained using RDKit, where the nodes represent atoms and the edges represent chemical bonds. The GIN has a dimension of 300 and 5 layers. The input molecular fingerprint is formed by concatenating Morgan, MACCS, and RDK fingerprints, resulting in a concatenated dimension of 2215, with Morgan and RDK fingerprints having dimensions of 1024 and MACCS having a dimension of 167. The concatenated features are processed by two layers of MLP, with the output dimension being 300. Then, the encoded molecular graph and fingerprint are concatenated, passed through two additional layers of MLP, and ultimately produce a molecular representation $\boldsymbol{h}_m$ with a dimension of 256.
In the spectrum encoder $g_s$, the number of chemical shift tokens for C-NMR is 2,500, while for H-NMR, it is 1,500. Additionally, the number of splitting pattern tokens is 75, and the number of coupling constant tokens is 200. The Transformer encoder~\cite{DBLP:conf/nips/VaswaniSPUJGKP17} used has a dimension of 256 and consists of 2 layers.

\paragraph{Training.}
We collect 9,000 molecules from ZINC~\cite{sterling2015zinc} and use simulation method~\cite{banfi2008nmrdb} to obtain the corresponding C-NMR and H-NMR data. Additionally, we gather extra 200 molecules to serve as the validation set. The training is conducted over 100 epochs, with hyperparameter $\tau$ set to 0.07.  The model checkpoint corresponding to the epoch with the lowest validation loss is saved.

\subsection{Inference Setting}
\label{sec:appendix-impl}

\begin{figure}[t]
    \centering
    \includegraphics[width=\linewidth]{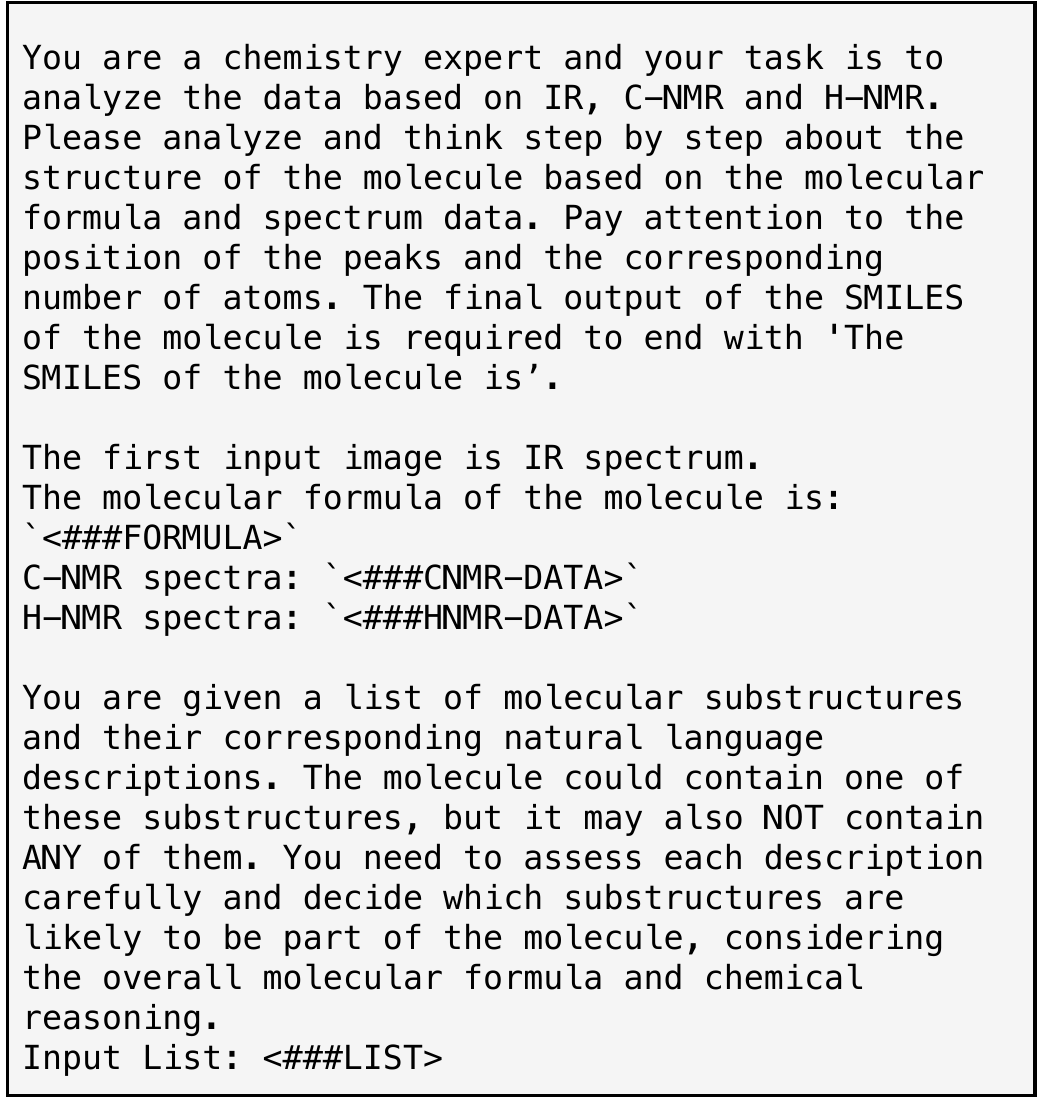} 
    \caption{Prompt for initialization.} 
    \label{fig:prompt-question} 
\end{figure}

\begin{figure}[t]
    \centering
    \includegraphics[width=\linewidth]{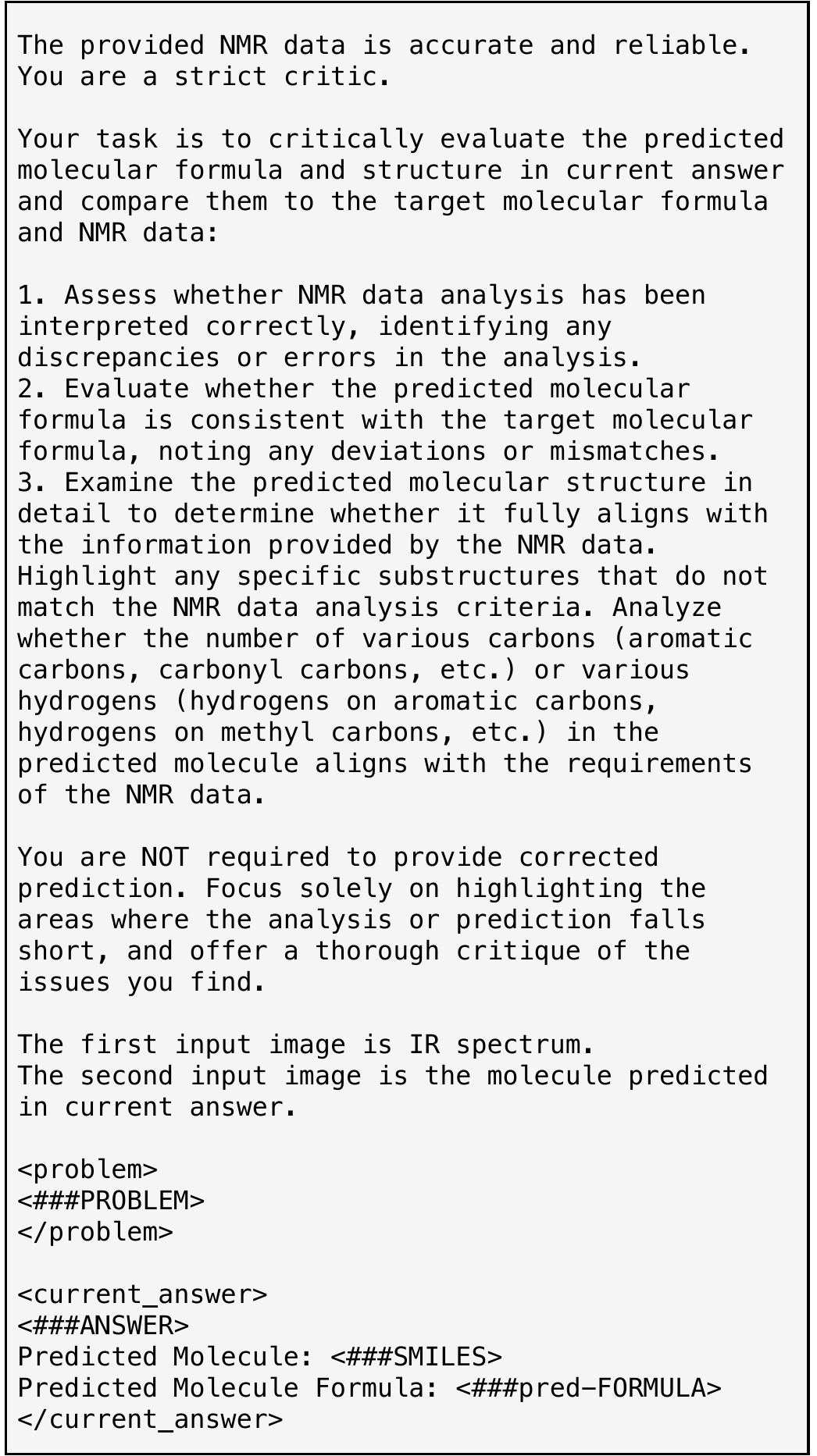} 
    \caption{Prompt for critique.} 
    \label{fig:prompt-critique} 
\end{figure}

\begin{figure}[t]
    \centering
    \includegraphics[width=\linewidth]{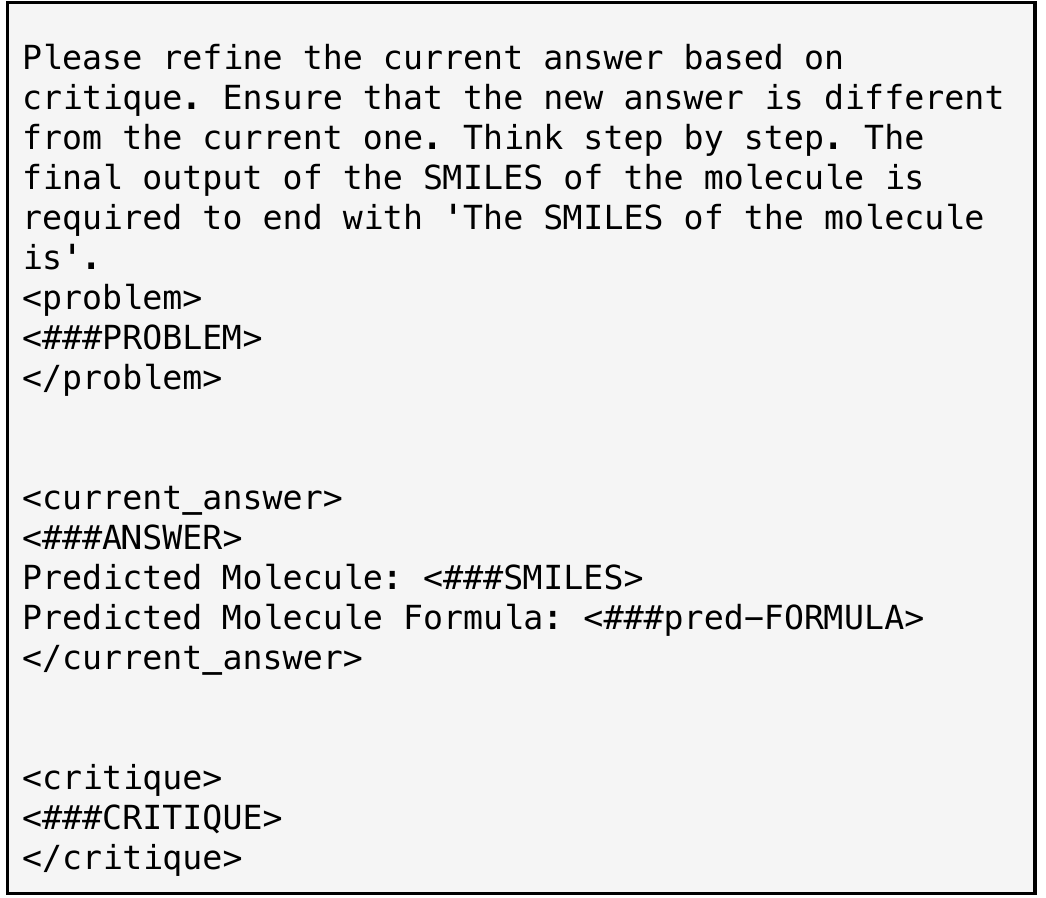} 
    \caption{Prompt for rewrite.} 
    \label{fig:prompt-refine} 
\end{figure}

During the inference phase, the parameters $c$ and $\epsilon$ in Equation~\eqref{eq:uct} of MCTS are set to 1 and 0.1, respectively, with a maximum of 2 child nodes per node. When the base models are Llama-3.2-11B, GPT-4o-mini, and GPT-4o, $N_{iter}$ is set to 8; when the base model is GPT-o1, $N_{iter}$ is set to 2. For Llama-3.2-11B and GPT-4o-mini, the number of retrieval count $k$ from the knowledge base is set to 1; while for GPT-4o and GPT-o1, $k$ is set to 2.
To maximize reproducibility, we set the temperature to 0.0 (most conservative) and top\_p to 1.0.
The prompts used during inference are shown in Figures~\ref{fig:prompt-question}, \ref{fig:prompt-critique} and \ref{fig:prompt-refine}.

\begin{table*}[t]
\centering
\scalebox{0.85}{
\begin{tabular}{lccccc}
\toprule
 & Morgan FTS & MACCS FTS & RDK FTS & Formula ACC & ACC \\ \midrule
CoT & 0.260 & 0.512 & 0.337 & 0.185 & 0.037 \\
CoT+$\mathcal{KB}$ & 0.337 \scalebox{0.9}{($+$0.077)} & 0.572 \scalebox{0.9}{($+$0.060)} & 0.413 \scalebox{0.9}{($+$0.076)} & 0.255 \scalebox{0.9}{($+$0.070)} & 0.111 \scalebox{0.9}{($+$0.074)} \\ \midrule
Self-R & 0.287 & 0.523 & 0.369 & 0.231 & 0.069 \\
Self-R+$\mathcal{KB}$ & 0.372 \scalebox{0.9}{($+$0.085)} & 0.598 \scalebox{0.9}{($+$0.075)} & 0.444 \scalebox{0.9}{($+$0.075)} & 0.356 \scalebox{0.9}{($+$0.125)} & 0.157 \scalebox{0.9}{($+$0.088)} \\ \midrule
Self-C & 0.299 & 0.535 & 0.373 & 0.222 & 0.083 \\
Self-C+$\mathcal{KB}$ & 0.398 \scalebox{0.9}{($+$0.099)} & 0.602 \scalebox{0.9}{($+$0.067)} & 0.458 \scalebox{0.9}{($+$0.085)} & 0.347 \scalebox{0.9}{($+$0.125)} & 0.171 \scalebox{0.9}{($+$0.088)} \\ \midrule
MAD & 0.292 & 0.520 & 0.359 & 0.222 & 0.079 \\
MAD+$\mathcal{KB}$ & 0.343 \scalebox{0.9}{($+$0.051)} & 0.563 \scalebox{0.9}{($+$0.043)} & 0.403 \scalebox{0.9}{($+$0.044)} & 0.273 \scalebox{0.9}{($+$0.051)} & 0.130 \scalebox{0.9}{($+$0.051)} \\ \midrule
MCTSr & 0.281 & 0.530 & 0.361 & 0.176 & 0.069 \\
MCTSr+$\mathcal{KB}$ & 0.333 \scalebox{0.9}{($+$0.052)} & 0.551 \scalebox{0.9}{($+$0.021)} & 0.414 \scalebox{0.9}{($+$0.053)} & 0.267 \scalebox{0.9}{($+$0.091)} & 0.106 \scalebox{0.9}{($+$0.037)} \\ \bottomrule
\end{tabular}
}
\caption{Performance of knowledge base integration on GPT-4o-mini across all methods. +$\mathcal{KB}$ denotes the incorporation of the knowledge base. The values in () indicate the relative performance improvement compared to when the knowledge base is not used.}
\label{tab:kb}
\end{table*}

\section{Baselines}
\label{sec:appendix-baseline}
\paragraph{CoT.} Chain-of-Thought (CoT)~\cite{DBLP:conf/nips/Wei0SBIXCLZ22} enhances the reasoning capabilities of LLMs by prompting them with ``\textit{think step by step}'' and generating intermediate reasoning steps to tackle complex tasks.

\paragraph{Self-Refine.} Self-Refine (Self-R)~\cite{DBLP:conf/nips/MadaanTGHGW0DPY23} iteratively improves the outputs of LLMs by generating feedback and refining its previous outputs without requiring additional training or supervised data. In the experiments, the number of refinement iterations is also set to \( N_{\text{iter}} \).

\paragraph{Self-Consistency.} Self-Consistency (Self-C)~\cite{DBLP:conf/iclr/0002WSLCNCZ23} improves reasoning in LLMs by sampling multiple reasoning paths and selecting the most consistent answer. In the experiments, the number of reasoning paths is also set to \( N_{\text{iter}} \).
\paragraph{MAD.}  The Multi-Agent Debate (MAD)~\cite{DBLP:conf/emnlp/Liang0JW00Y0T24} encourages divergent thinking in LLMs by involving multiple agents in a debate process, where they express their arguments and a judge manages the debate to reach a final solution. In the experiments, the maximum number of debate rounds is 4.

\paragraph{MCTSr.} MCT Self-Refine (MCTSr)~\cite{DBLP:journals/corr/abs-2406-07394} integrates LLMs with Monte Carlo Tree Search (MCTS) to enhance performance in complex reasoning tasks by iteratively refining solutions through systematic exploration and heuristic self-evaluation mechanisms. In the experiments, the number of iterations is also set to $N_{iter}$.

\begin{table*}[t]
\centering
\scalebox{0.85}{
\begin{tabular}{lccccc}
\toprule
 & Morgan FTS & MACCS FTS & RDK FTS & Formula ACC & ACC \\ \midrule
K-MSE & 0.470 & 0.651 & 0.520 & 0.412 & 0.273 \\ \midrule
w/o Fml & 0.432 & 0.627 & 0.474 & 0.347 & 0.218 \\
w/o Img & 0.451 & 0.628 & 0.502 & 0.380 & 0.236 \\
w/o Fml, w/o Img & 0.427 & 0.620 & 0.469 & 0.324 & 0.208 \\ \bottomrule
\end{tabular}
}
\caption{Experimental results of critique input ablation study.}
\label{tab:critique-input}

\end{table*}

\section{Additional Experimental Results}
\label{sec:appendix-results}

\subsection{Universal Performance Gains through Knowledge Base Integration}
LLMs often face the challenge of insufficient knowledge when handling tasks in the chemical domain. To address this, the proposed molecular substructure knowledge base can be used as a plugin,  integrated into any LLM's reasoning framework. We conduct experiments by integrating this knowledge base with all baseline reasoning methods, and the results are shown in Table~\ref{tab:kb}. The findings indicate that incorporating the knowledge base consistently improves model performance, validating its effectiveness in enhancing reasoning capabilities. This also demonstrates that the knowledge base is highly adaptable and flexible, capable of functioning across various reasoning frameworks.
\subsection{Ablations on Critique Input}
The key to iterative self-reflection and improvement lies in the model's ability to identify and correct defects in the previous reasoning step. However, we find that due to the limited capability of LLMs in the field of chemical molecules, they often struggle to accurately analyze the molecular structure of the molecule predicted in the previous step. Specifically, when analyzing the predicted molecule represented in SMILES format, the model frequently has difficulty correctly interpreting the corresponding molecular structure during the self-critique process, making the critique unreliable and weakening the effectiveness of self-reflection and improvement.

To address this issue, as shown in Equation~\eqref{eq:critique}, we introduce molecular formula and image of the predicted molecule during the critique process, both of which can be generated using RDKit. This additional information helps LLMs more accurately identify the molecular structure, thereby improving the reliability of the critique and enhancing the effectiveness of self-reflection.

We further conduct ablation experiments, and the results are shown in Table~\ref{tab:critique-input}, where w/o Img and w/o Fml refer to settings in which the molecular image and molecular formula are removed during the critique process, respectively. The results indicate that removing either image or formula leads to a decline in model performance, suggesting that these supplementary inputs play a crucial role in the critique process. They can improve the model’s ability to recognize molecular structures, thereby ensuring the effectiveness of the critique and self-improvement mechanism.

\begin{table*}[t]
\centering
\scalebox{0.85}{
\begin{tabular}{ll}

\toprule
\multicolumn{2}{l}{\textit{Question}} \\ \midrule
   Infrared spectroscopy data ($x_{\text{ir}}$)       &    \raisebox{-0.5\height}{\includegraphics[width=0.5\linewidth]{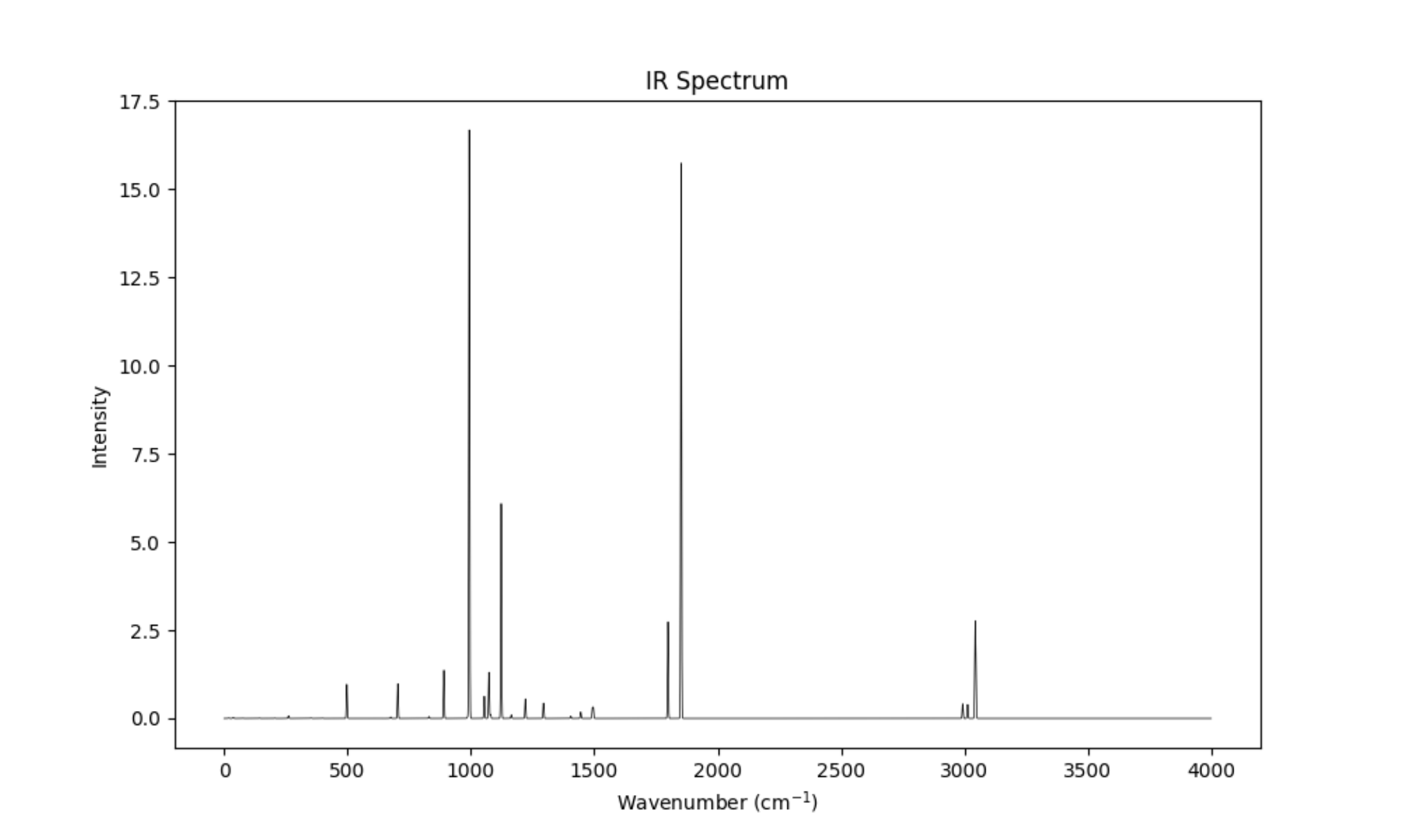} }     \\
    C-NMR data ($x_{\text{cnmr}}$)     &    \texttt{9.1 (2C, s), 27.8 (2C, s), 170.5 (2C, s)}      \\
     H-NMR data  ($x_{\text{hnmr}}$)    &    \texttt{1.06 (6H, t,  J  = 7.3 Hz), 2.50 (4H, q,  J  = 7.3 Hz)}      \\ 
     Molecular formula  ($x_{\text{formula}}$)    &         \texttt{C6H10O3} \\\midrule
\multicolumn{2}{l}{\textit{Target}} \\ \midrule
   molecule in SMILES     &    \texttt{CCC(=O)OC(=O)CC}      \\ \bottomrule
\end{tabular}
}
\caption{Example input and target data (ID: 5-43).}
\label{tab:example}
\end{table*}

\subsection{Ablations on the Component of Specialized Scorer}

\begin{figure}[t]
    \centering
    \includegraphics[width=0.8\linewidth]{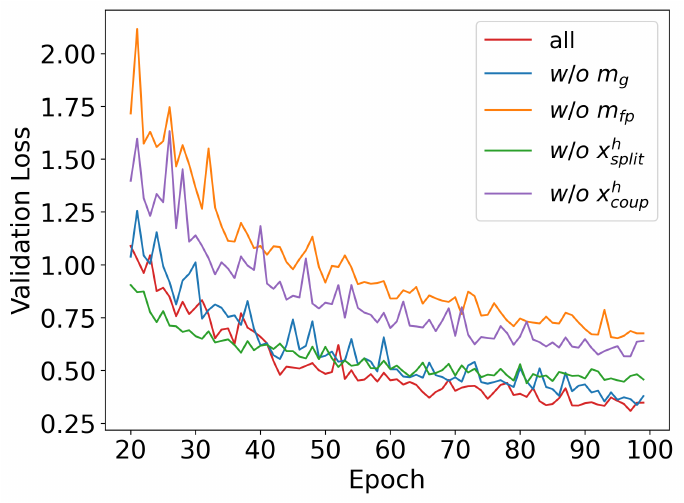} 
    \caption{Visualization of validation loss during the scorer training process.} 
    \label{fig:scorer-loss} 
\end{figure}

We analyze the specialized molecule-spectrum scorer through ablation experiments. The variants in the experiments include: w/o $m_g$ (removing molecular graph input in the molecule encoder $g_m$); w/o $m_{fp}$ (removing molecular fingerprint input in the molecule encoder $g_m$); w/o $x^h_{\text{split}}$ (removing hydrogen atom splitting pattern input in the spectrum encoder $g_s$); and w/o $x^h_{\text{coup}}$ (removing hydrogen atom coupling pattern input in the spectrum encoder $g_s$). Figure~\ref{fig:scorer-loss} shows the visualization of the validation loss during the training process for these variants, where ``all'' represents the complete scorer. From the results, it is observed that removing any component causes an increase in the validation loss after training convergence. This indicates that each component plays a role in capturing the relationship between the molecule and the spectrum data, and the removal of these components affects the model's performance.

\begin{figure}[t]
    \centering
    \includegraphics[width=1\linewidth]{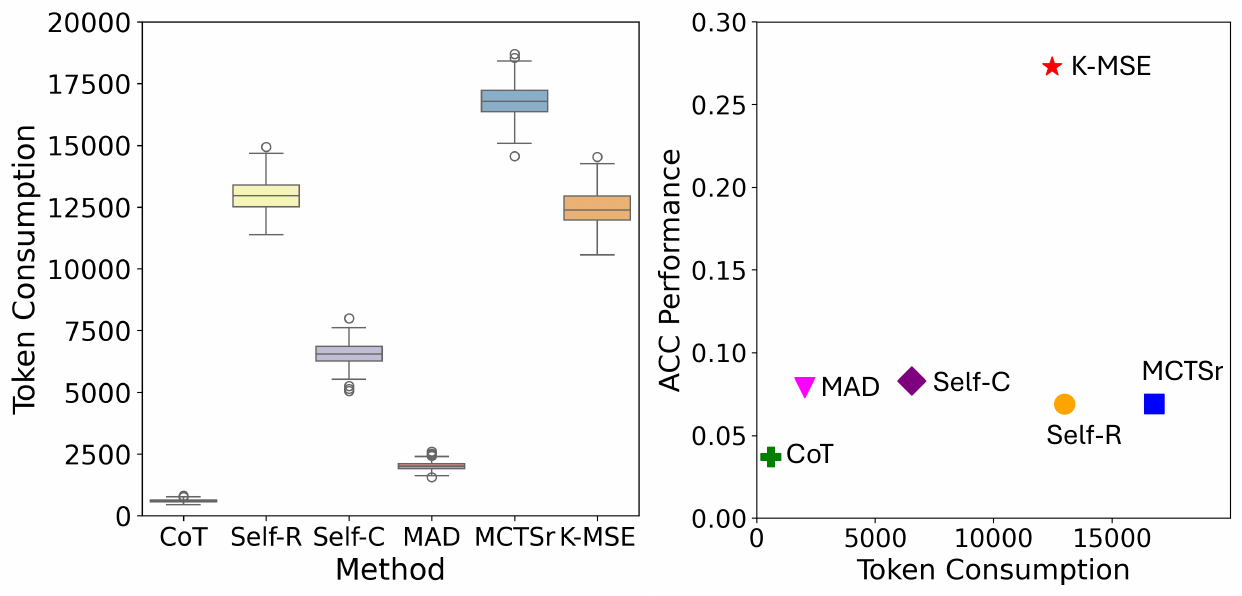} 
    \caption{Left: token consumption per \textit{question} for each method. Right: ACC performance versus token consumption across different methods.} 
    \label{fig:token} 
\end{figure}
\subsection{Token Consumption Analysis}
In Figure~\ref{fig:token}, we analyze the token consumption on GPT-4o-mini of different methods when solving each problem (left) and the relationship between token consumption and performance (right). K-MSE maintains a relatively reasonable token consumption level, which is lower than that of the Self-R and MCTSr methods. Notably, K-MSE, as an efficient test-time scaling method, effectively utilizes token consumption during test-time to significantly enhance model performance.
We attribute this to two key factors: first, the effective expansion of the LLM's knowledge capabilities through the introduction of a molecular substructure knowledge base; second, the development of a specialized scorer for evaluating reasoning outcomes and enhancing self-critique, combined with additional inputs such as molecular images and formulas, enables LLMs to reflect and improve.
In contrast, while some methods increase token consumption to acquire more reasoning steps, their trade-off between resource consumption and performance improvement is not optimal.

\section{Illustrative Data Example}
\label{sec:example}
We provide an illustrative data example (ID 5-43 in the dataset) in Table~\ref{tab:example} to illustrate the task's input-output specifications.
\clearpage
\end{document}